\documentclass[11pt,a4paper]{article}
\usepackage[margin=25mm,top=23mm,bottom=23mm]{geometry}
\usepackage[T1]{fontenc}
\usepackage{lmodern}
\usepackage{microtype}
\usepackage{amsmath,amssymb,amsthm}
\usepackage{graphicx}
\usepackage{booktabs,tabularx,multirow,array}
\usepackage[dvipsnames]{xcolor}
\usepackage{float}
\usepackage{caption}
\usepackage{subcaption}
\usepackage{tcolorbox}
\tcbuselibrary{skins,breakable}
\usepackage{enumitem}
\usepackage{titlesec}
\usepackage{fancyhdr}
\usepackage[round,authoryear]{natbib}
\usepackage{hyperref}
\usepackage{url}
\usepackage{siunitx}
\usepackage{longtable}
\graphicspath{{figures/}}
\definecolor{navy}{HTML}{143958}
\definecolor{teal}{HTML}{087E86}
\definecolor{ochre}{HTML}{B67A30}
\definecolor{muted}{HTML}{637A89}
\definecolor{paleblue}{HTML}{EFF5F8}
\definecolor{figpink}{HTML}{E8B5C5}
\definecolor{figblue}{HTML}{A8C8E8}
\hypersetup{colorlinks=true,linkcolor=navy,citecolor=teal,urlcolor=teal,pdftitle={AutoAlphaBench: Can an Agent Find Alpha that Survives the Future?},pdfauthor={Kangcheng Deng, Qixuan Yang, Hui Cai, Jiacheng Lu, Xianwei Li, Rui Sun, Beidi Luan, Jing Li, Zuo Bai}}
\titleformat{\section}{\normalfont\Large\bfseries\color{navy}}{\thesection}{0.65em}{}
\titleformat{\subsection}{\normalfont\large\bfseries\color{navy}}{\thesubsection}{0.65em}{}
\titleformat{\subsubsection}{\normalfont\normalsize\bfseries\color{teal}}{\thesubsubsection}{0.65em}{}
\titlespacing*{\section}{0pt}{1.55em}{.65em}
\titlespacing*{\subsection}{0pt}{1.08em}{.37em}
\setlist[itemize]{topsep=3pt,itemsep=2pt,parsep=0pt,leftmargin=1.3em}
\setlist[enumerate]{topsep=3pt,itemsep=2pt,parsep=0pt,leftmargin=1.5em}
\newtcolorbox{finding}[1][]{enhanced,breakable,colback=paleblue,colframe=white,boxrule=0pt,borderline west={3pt}{0pt}{teal},arc=1pt,left=9pt,right=8pt,top=6pt,bottom=6pt,#1}
\newcommand{\obs}[1]{\textbf{\color{teal}Observation #1.}}

\newcommand{\figfile}[2][.99\linewidth]{\includegraphics[width=#1]{#2}}
\newcommand{\NA}{\textemdash}
\newcommand{\betterup}{\ensuremath{\,\uparrow}}

\begin{document}
\begin{center}
{\Huge\sffamily\bfseries AutoAlphaBench:}\par\vspace{3pt}
{\Large\bfseries Can an Agent Find Alpha that Survives the Future?}\par\vspace{9pt}
{\normalsize
Kangcheng Deng\textsuperscript{1,2,*},\quad
Qixuan Yang\textsuperscript{1,3,*},\quad
Hui Cai\textsuperscript{1},\quad
Jiacheng Lu\textsuperscript{1,3},\quad
Xianwei Li\textsuperscript{1}\par
Rui Sun\textsuperscript{1},\quad
Beidi Luan\textsuperscript{1},\quad
Jing Li\textsuperscript{1},\quad
Daxin Jiang\textsuperscript{1},\quad
Zuo Bai\textsuperscript{1,4,\dag}\par}
{\small
\textsuperscript{1}StepFun\qquad
\textsuperscript{2}Fudan University\qquad
\textsuperscript{3}Shanghai Jiao Tong University\qquad
\textsuperscript{4}FinStep\par}
{\footnotesize
\textsuperscript{*}Equal contribution\qquad
\textsuperscript{\dag}Corresponding author\par}
\end{center}
\begin{tcolorbox}[colback=paleblue,colframe=paleblue,boxrule=0pt,arc=2mm,left=13pt,right=13pt,top=7pt,bottom=6pt]
\vspace{8pt}
\textbf{Abstract.} Quantitative investment research seeks signals that hold up on new market data and add distinct sources of return to a portfolio. Language-model agents can implement and backtest a human given hypotheses, yet existing evaluations offer limited evidence on whether they can successfully choose what deserves investigating. We thus introduce \textbf{AutoAlphaBench} to compare four research settings under a common evaluation protocol. Agents \emph{investigate a provided hypothesis} to test execution capability, select a hypothesis within an assigned research direction in \emph{Guided}, and choose what to investigate independently in \emph{Free} or with access to earlier proposals in \emph{Cascaded}. We assess out-of-sample factor quality, factor diversity and portfolio performance. Experiments show that model rankings shifted under independent hypothesis selection, with Opus 5.5 falling from first in \emph{Provided Hypothesis} to last in \emph{Free}. Compared with \emph{Guided}, \emph{Cascaded} produced greater factor diversity and lower mean factor quality. Its portfolios had smaller maximum drawdowns in most matched comparisons, with lower annualized excess returns on average. These findings motivate research on how autonomous agents can build on prior proposals to discover more complementary signals while sustaining out-of-sample quality.

\vspace{6pt}

\vspace{6pt}
{\color{teal}\small\bfseries Homepage: \url{https://auto-alpha.trade}}\par
\vspace{6pt}

\end{tcolorbox}
\begin{figure}[H]
\centering
\figfile[.9\linewidth]{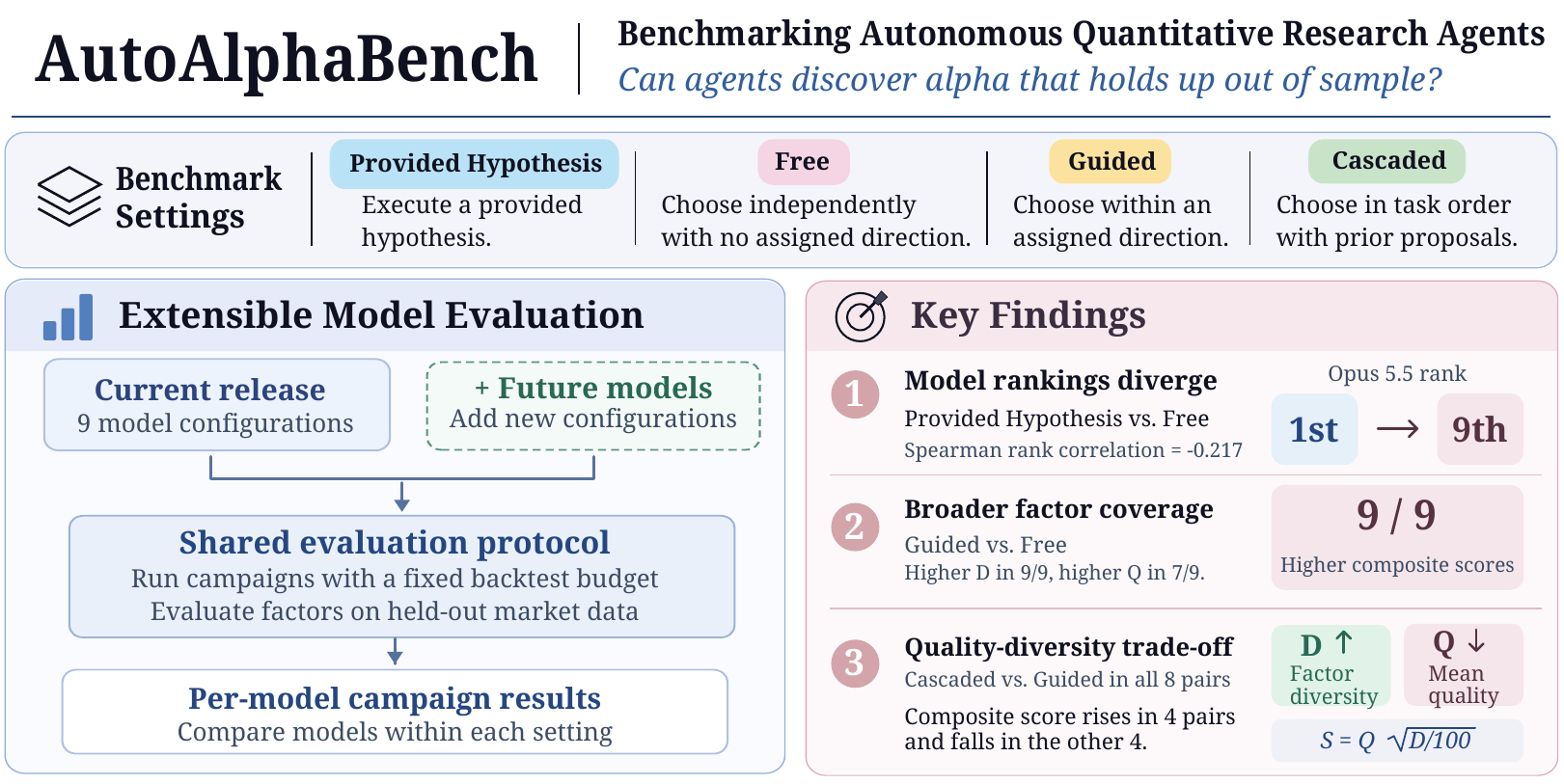}
\caption{\textbf{AutoAlphaBench evaluates research agents in quantitative factor discovery.} Provided Hypothesis evaluates how well agents investigate supplied hypotheses. In Free, agents independently choose hypotheses without an assigned research direction. In Guided, they choose hypotheses within assigned directions. In Cascaded, they choose hypotheses in task order with access to earlier proposals. The three autonomous settings report factor diversity and a composite score that combines mean factor quality with diversity, which can be used to access additional models.}
\label{fig:teaser}
\end{figure}

\section{Introduction}

Large language model (LLM) agents can propose hypotheses, implement them, and revise experiments in response to feedback \citep{lu2026aiscientist,li2025rdagentquant}. When agents run a series of investigations under a limited budget, they must decide both \emph{which hypotheses to investigate} and \emph{how to test them}. An agent that performs well on assigned problems can still spend much of its budget testing hypotheses that yield weak or redundant results.

Existing research agent benchmarks evaluate how well agents carry out assigned tasks, such as running ML experiments, replicating papers, or solving problems derived from published studies \citep{huang2024mlagentbench,wijk2025rebench,starace2025paperbench,researchclawbench}. Scientific ideation benchmarks evaluate the proposals themselves, including their novelty, feasibility, and grounding in prior work \citep{si2025ideas,agentideabench}. In quantitative finance, benchmarks assess generated factors and executable trading strategies \citep{alphabench,alphaforge,quantarena}. Report2Alpha-Bench provides brokerage reports as starting points for agents to formulate factor hypotheses and refine them through repeated backtesting \citep{report2alpha}. However, these benchmarks leave open how performance differs between assigned and self-selected hypotheses.

To address this gap, we introduce \textbf{AutoAlphaBench}, a benchmark for repeated investigations in quantitative factor discovery. In \emph{Provided Hypothesis}, agents investigate a hypothesis supplied by a researcher, reflecting a workflow centered on validating a candidate signal. \emph{Cascaded} represents the goal of sustained autonomous factor discovery. Agents choose successive hypotheses with access to earlier proposals, allowing us to assess whether they can sustain productive and diverse discovery with minimal human guidance. To examine how agents generate and investigate their own hypotheses under different levels of guidance, we include two further settings. In \emph{Free}, agents choose hypotheses independently without an assigned research direction. In \emph{Guided}, they choose within assigned directions, allowing us to assess whether directional guidance helps them identify more promising hypotheses and explore a broader range of ideas across the campaign. In all four settings, agents develop and submit executable factors under a specified backtest budget, and we report mean held-out factor quality, factor diversity, and a composite score for model ranking.


Experiments show that under the same backtest budget, rankings on provided hypotheses carry over poorly when agents decide which hypotheses to pursue, with Opus 5.5 falling from first in \emph{Provided Hypothesis} to last in \emph{Free}. This contrast motivates us to examine how providing and structuring information can help agents decide which research hypotheses to pursue. Specifically, assigning research directions in \emph{Guided} yields higher composite scores and greater factor diversity than \emph{Free} for all nine model configurations. In \emph{Cascaded}, by consulting earlier proposals, agents generate more diverse discoveries but have lower mean held-out quality, revealing autonomous discoveries converge on less productive regions of the hypothesis space. Portfolio level evidence further shows that the greater factor diversity in \emph{Cascaded} is accompanied by smaller maximum drawdowns in most matched comparisons with \emph{Guided}, alongside lower annualized excess returns on average. Together, these findings suggest that future autonomous research work should focus on how agents allocate a limited research budget to discover factors that hold up out of sample and complement one another in a portfolio. Our contributions are threefold:
\begin{itemize}

    \item \textbf{A benchmark for autonomous factor discovery.}
    We introduce AutoAlphaBench to evaluate agents as responsibility for hypothesis selection shifts from researchers to agents. 
    
    \item \textbf{Cross-setting evidence on model capability.}
    We find rankings on supplied hypotheses carry over poorly to independent hypothesis selection, revealing that assessments of model capability depend on who selects the hypotheses and how the search is structured.

    \item \textbf{Portfolio implications of the quality--diversity trade-off.}
    We examine what broader factor discovery contributes once the submitted factors are combined into portfolios.

\end{itemize}

\section{Related Work}
\textbf{Scientific ideation and research agents.}
SciMON retrieves prior work to ground idea generation and uses the retrieved literature to revise proposals for novelty \citep{wang2024scimon}. \citet{si2025ideas} compare LLM-generated and human-written proposals through blinded expert review and report limited diversity in their ideation agent. AgentIdeaBench evaluates literature-grounded proposal generation and revision \citep{agentideabench}.
The AI Scientist integrates idea generation, experimentation, and manuscript preparation \citep{lu2026aiscientist}. PaperBench evaluates replication of assigned papers, RE-Bench evaluates open-ended ML engineering tasks, and ResearchClawBench derives tasks from published studies while withholding the target paper \citep{starace2025paperbench,wijk2025rebench,researchclawbench}. AutoAlphaBench varies the information available when agents choose hypotheses and evaluates the held-out outcomes of the implementations produced across repeated investigations.

\textbf{Automated factor mining.}
AlphaGen uses reinforcement learning to search for sets of formulas that work together \citep{yu2023alphagen}. Alpha-GPT supports interactive factor mining with human input, while R\&D-Agent-Quant automates joint factor and model optimization \citep{wang2025alphagpt,li2025rdagentquant}.
AlphaAgent regularizes exploration, QuantaAlpha evolves mining trajectories, and FactorMiner stores reusable skills and experiences across mining iterations \citep{alphaagent,quantaalpha,wang2026factorminer}. FactorMiner makes experience retention part of its mining algorithm. In AutoAlphaBench, access to prior proposals is part of the Cascaded benchmark condition, and the evaluation tracks both held-out quality and diversity across submitted factors.

\textbf{Benchmarks of quantitative research and trading.}
AlphaBench evaluates formulaic factor generation, evaluation, and search \citep{alphabench}. AlphaForgeBench evaluates end-to-end trading strategy design, and Quant-Arena tests executable strategies on a frozen out-of-sample interval \citep{alphaforge,quantarena}.
Report2Alpha-Bench is particularly close to our setting. It supplies brokerage research reports, lets agents propose and refine hypotheses over multiple backtesting rounds, and analyzes within-round return correlations and held-out factor-pool performance \citep{report2alpha}. AutoAlphaBench compares a setting with supplied hypotheses and three settings in which agents select their own hypotheses. The evaluator reports mean task quality and factor diversity for campaigns in the autonomous settings.

\textbf{Multiple testing and factor diversity.}
Repeated searches over candidate signals complicate the interpretation of factor significance and backtest performance \citep{harvey2016factorzoo,bailey}. We use a sealed period to assess the submitted factors and a return-based diversity statistic to characterize the submissions within each campaign. Scientific originality relative to the published literature is a separate question.

\section{Benchmarking Autonomous Research}
\subsection{Task Formulation}
Let $\mathcal D_{\mathrm{dev}}$ denote the market data available for development and $\mathcal D_{\mathrm{test}}$ the data from the held-out assessment period. A research proposal $z_i$ describes an economic mechanism and a candidate signal. The executor implements the proposal as a factor $f_i$ with parameters $\theta_i$. During development periods, an agent can query a bounded backtest tool and observe feedback $u_{i,1:B}$. The backtest time limit in the reported evaluation is $B\leq 5$. The agent submits one executable factor before test evaluation, as follows.
\begin{equation}
 z_i\sim\pi_m(\cdot\mid o_i,h_{<i}),\quad
 f_i\leftarrow\mathcal E_m(z_i,\mathcal D_{\rm dev},u_{i,1:B}),\quad
 q_i=\mathcal V(f_i;\mathcal D_{\rm test}).\label{eq:task}
\end{equation}
Here $o_i$ is the condition-specific task instruction, and $h_{<i}$ denotes accessible prior research information. In Provided Hypothesis, the instruction supplies $z_i$. In the other conditions, the agent selects $z_i$ under the information specified by the protocol. Equation~\ref{eq:task} represents proposal selection and implementation as separate steps in the workflow. Individual submissions are subject to delivery and audit checks. For autonomous settings, the evaluator reports both mean task quality $Q_m$ and factor diversity $D_m$ for each campaign $\mathcal C_m=\{(z_i,f_i,q_i)\}_{i=1}^{N_m}$.

\subsection{Benchmark Design}

\textbf{Task structure.} Each task specifies (i) a supplied hypothesis or instructions for selecting one, (ii) access to development market data and a backtest tool with budget, and (iii) a requirement to submit one executable factor. The agent uses development backtest feedback to refine the factor before submission. The evaluator then assesses submitted factor on held-out market data.

\textbf{Research settings.} In \emph{Provided Hypothesis}, the benchmark supplies the hypothesis and the agent decides how to investigate it. In \emph{Free}, the agent independently selects a hypothesis for each task without an assigned direction. In \emph{Guided}, the agent selects a hypothesis within an assigned research direction. The reported evaluation assigns ten tasks to each of ten predefined directions. In \emph{Cascaded}, tasks run in order and the agent can consult earlier proposals before selecting the next hypothesis. Task instructions, assignment, order, and access to prior proposals can vary together, so we report comparisons between settings as implemented.

\textbf{Evaluation protocol and released records.} Additional model configurations can be evaluated using the same task instructions, backtest budget, held-out market data, and scoring rules within each setting. The release includes submitted code or factors, proposal names where available, audit and delivery status, development and test period Sharpe ratios, backtest counts, and selected trajectory events. For autonomous campaigns, it also reports the evaluator's effective-rank-based measure of factor-return diversity. Recomputing this measure requires the underlying stock-by-day return matrices.


\begin{figure}[t]
\centering
\figfile[.98\linewidth]{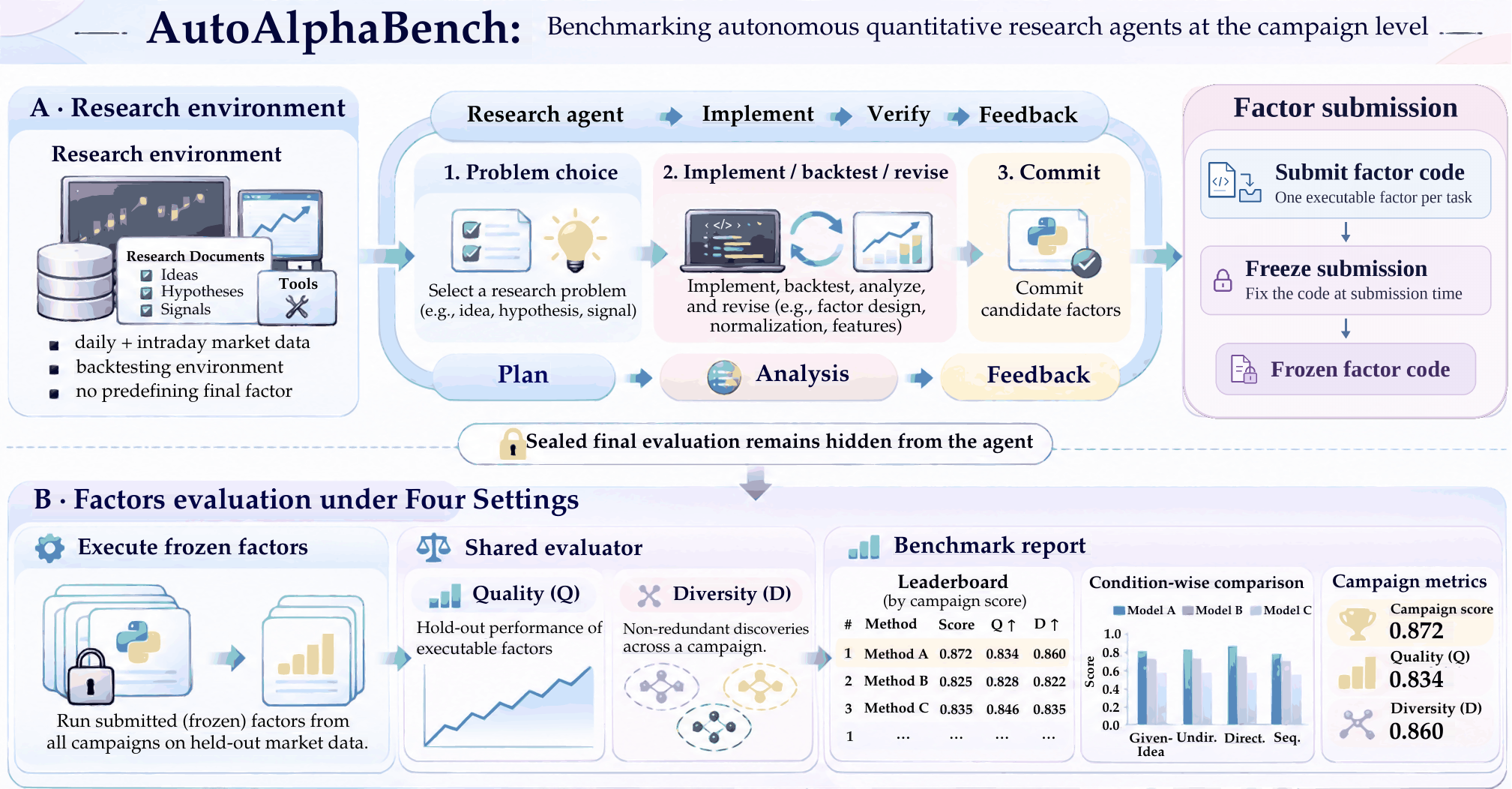}

\caption{\textbf{AutoAlphaBench research and evaluation workflow.}
\textbf{(A) Factor development and submission.} Under the assigned research condition, the agent investigates a hypothesis, implements a candidate factor, and revises it using feedback from development-period backtests. The agent submits one executable factor per task, and the submitted code is frozen before evaluation.
\textbf{(B) Factors evaluation under Four Settings.} A shared evaluator runs the frozen factors on held-out market data, scores individual submissions, and measures diversity among factors. The held-out data and evaluation results are withheld from the agent
during research.}
\label{fig:design}
\end{figure}

\subsection{Evaluation Protocol}\label{sec:protocol}
For each campaign in Free, Guided, or Cascaded, the evaluator aggregates task quality and factor diversity as follows.
\begin{equation}
 Q_m=\frac{1}{N_m}\sum_{i=1}^{N_m}q_i,\qquad
 S_m=Q_m\sqrt{D_m/100} .\label{eq:score}
\end{equation}
The reported $q_i$ incorporates task-level assessment and validity rules, and $D_m$ is taken from the evaluator output. In Provided Hypothesis, the reported score is the mean of the per-task scores for the supplied hypotheses. Comparisons between Provided Hypothesis and the autonomous settings use within-setting ranks based on the score defined for each setting.

For the autonomous settings, we report $Q$ and $D$ alongside the fraction of zero-quality outcomes, delivery and audit status, development- and test-period Sharpe ratios, proposal-name concentration, and selected trajectory events. Proposal-name concentration is defined as $\mathrm{HHI}=\sum_k(n_k/N)^2$, where $n_k$ counts the proposals whose normalized \texttt{idea.name} string is $k$ among $N$ proposals. This statistic measures exact-name repetition only. Cross-model analyses are exploratory, because they compare a small set of model configurations whose tasks can share market periods.

\section{Benchmark Results}

\label{sec:results}
\subsection{Experimental Setup}
We analyze the model-information settings under the four benchmark conditions. The protocol admits additional model configurations without changing the task interface or evaluation rules. Appendix~\ref{app:configuration} reports the observed denominators for the present release. The benchmark specifies a nominal limit of five backtests per task, and the exported records include occasional exceptions to that limit. Financial outcomes are taken from the evaluator output.

\textbf{Outcome measures.} Provided Hypothesis reports the mean task score $G$ for supplied hypotheses. The autonomous settings report mean task quality $Q$, campaign diversity $D$, and the composite score $S=Q\sqrt{D/100}$. Since $G$ and $S$ follow different scoring definitions, comparisons between Provided Hypothesis and the autonomous settings use within-setting ranks. Comparisons among Free, Guided, and Cascaded use paired changes in $Q$, $D$, and $S$. We additionally report portfolio results for investment reference.

\subsection{Main Results}

\label{sec:mainresults}
\begin{table}[!htbp]\centering\scriptsize
\caption{\textbf{Setting-specific benchmark outcomes.} Provided Hypothesis reports mean task score $G$. Each autonomous entry reports composite score $S$ followed by its components $(Q,D)$, where $Q$ is mean task quality and $D$ is campaign diversity. $G$ and $S$ use different scoring definitions, so their raw values are not directly comparable. Appendix~\ref{app:extended-diagnostics} reports zero-quality counts.}\label{tab:scores}
\begingroup
\setlength{\tabcolsep}{5.5pt}
\renewcommand{\arraystretch}{1.18}
\begin{tabular}{@{}lcccc@{}}\toprule
 & Provided Hypothesis & Free & Guided & Cascaded \\
Model & $G\betterup$ & $S\;(Q,D)\betterup$ & $S\;(Q,D)\betterup$ & $S\;(Q,D)\betterup$ \\ \midrule
Opus 5.5 & \textbf{24.57} & 1.45 (6.21, 5.44) & 14.01 (29.95, 21.88) & \textbf{16.66} (\textbf{21.70}, 58.90) \\
GPT-6.1 Sol & 23.48 & 5.10 (16.70, 9.31) & 13.11 (29.20, 20.17) & 16.14 (19.54, 68.24) \\
GPT-5.6 Sol & 20.17 & 2.00 (5.58, 12.80) & 11.43 (21.18, 29.12) & 14.72 (17.39, 71.65) \\
GPT-6 Astra & 22.35 & 2.99 (11.02, 7.36) & 12.52 (32.27, 15.05) & 13.83 (16.25, 72.47) \\
GLM 5.3 & 21.55 & 10.05 (21.77, 21.29) & \textbf{16.70} (32.16, 26.96) & 14.04 (16.41, \textbf{73.15}) \\
DeepSeek V4.1 & 22.27 & 12.06 (33.86, 12.68) & 14.23 (30.06, 22.40) & 15.48 (20.45, 57.30) \\
Kimi K3.1 & 23.43 & 7.01 (15.68, 20.00) & 13.70 (26.73, 26.27) & 13.60 (18.74, 52.62) \\
Step 5 Preview & 23.13 & \textbf{13.91} (\textbf{34.37}, 16.38) & 15.57 (\textbf{32.53}, 22.92) & 14.37 (17.99, 63.85) \\
HY4 & 23.66 & 8.90 (24.31, 13.39) & 13.83 (24.35, \textbf{32.25}) & 13.35 (20.57, 42.09) \\ \bottomrule
\end{tabular}
\endgroup
\par
\vskip 3pt
\end{table}

Table~\ref{tab:scores} reports the outcome profile for each setting. Together, the results show that \textbf{research performance depends on how investigation opportunities are allocated.} Performance on provided hypotheses is weakly related to performance when agents choose hypotheses, guided structure changes joint quality-diversity outcome, and cascaded research expands diversity while reducing mean task quality. The three observations below examine these distinctions in turn.

\begin{finding}\obs{1} \textbf{Provided Hypothesis and Free produce substantially different model rankings, separating supplied-hypothesis execution ability from autonomous hypothesis selection ability.}\end{finding}

\begin{figure}[!htbp]\centering
\figfile[.93\linewidth]{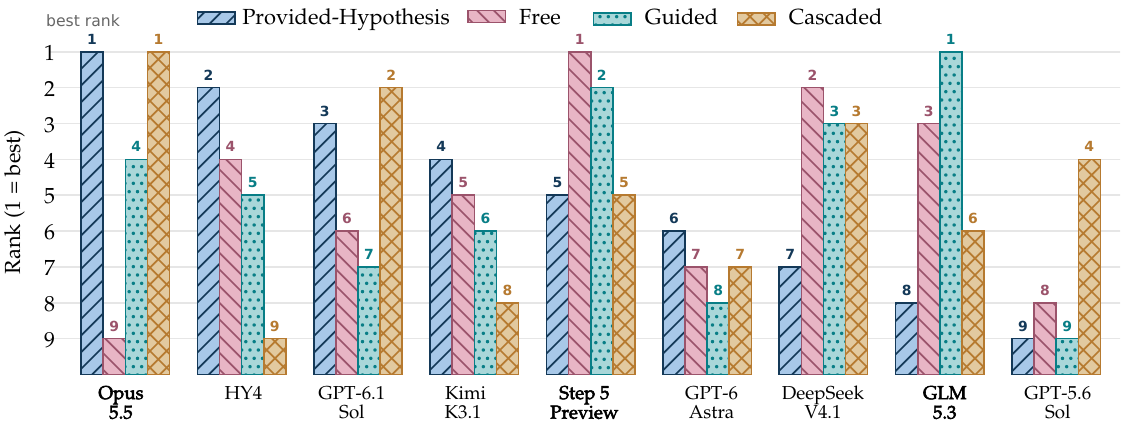}
\caption{\textbf{Rank comparison under supplied and self-selected hypotheses.} Grouped bars compare each model's within-setting rank in Provided Hypothesis and Free; taller bars indicate better ranks, and the number above each bar gives the rank. Models are ordered by Provided Hypothesis rank. The figure visualizes rank reordering, while Table~\ref{tab:scores} reports the underlying scores and autonomous score components.}\label{fig:scores}
\end{figure}

Opus 5.5 moves from first in Provided Hypothesis to last in Free, while Step 5 Preview moves from fifth to first, as shown in Figure~\ref{fig:scores}. The Spearman rank correlation between the two settings is $-0.217$. The change is linked to research allocation in the observed campaigns. In Free, three quarters of Opus 5.5 proposals concern liquidity and 63 submissions receive a quality score of zero, although all submissions pass the recorded delivery and audit checks. The model-level proposal profiles are reported in Appendix~\ref{app:model-profiles}.

\begin{figure}[!htbp]\centering
\figfile[.68\linewidth]{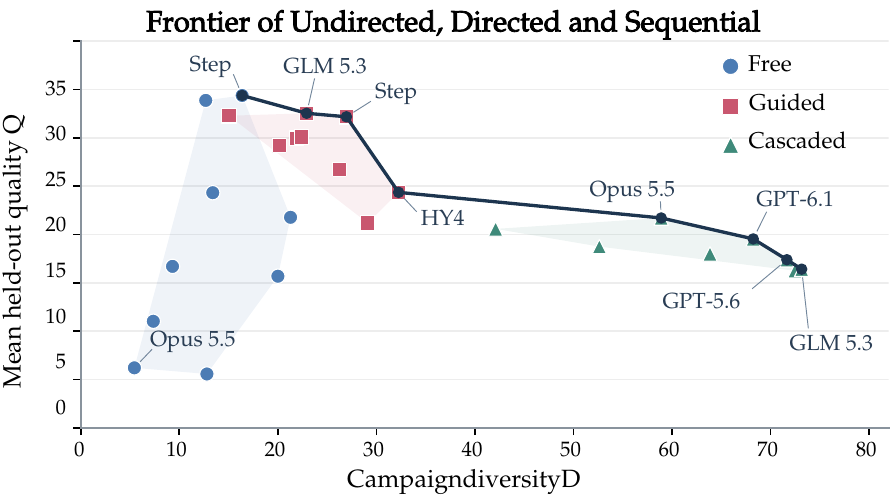}
\caption{\textbf{Free and Cascaded organization shift diversity and mean task quality differently.} Each point represents a model evaluated in both conditions of a paired comparison. Panel A reports Guided minus Free, and Panel B reports Cascaded minus Guided. Positive $\Delta D$ indicates higher campaign diversity in the second condition, while positive $\Delta Q$ indicates higher mean task quality. All points lie to the right of zero in both panels. Most points in Panel A lie above zero, whereas every point in Panel B lies below zero. The panels share a vertical scale and use separate horizontal scales.}\label{fig:guidance-deltas}
\end{figure}

\begin{finding}\obs{2} \textbf{Directional structure improves research allocation by broadening factor coverage while preserving or increasing mean task quality for most models.}\end{finding}

Relative to Free, Guided yields higher factor diversity in every paired campaign and higher mean task quality in most campaigns. Consequently, the composite score $S$, which combines quality and diversity, is higher under Guided in every paired comparison. Panel A of Figure~\ref{fig:guidance-deltas} shows this pattern directly: all points lie to the right of zero in diversity, and most lie above zero in mean quality. Three Guided campaigns also lie on the quality--diversity frontier formed by the Free and Guided campaigns. The portfolio results are consistent with this pattern, with higher annualized returns under Guided in all but one paired campaign. These results indicate that allocating investigations across predefined research directions is associated with broader factor-return coverage and generally stronger research outcomes. Because the two settings differ in their task-allocation rules, this comparison remains descriptive.

\begin{figure}[!htbp]\centering
\figfile[\linewidth]{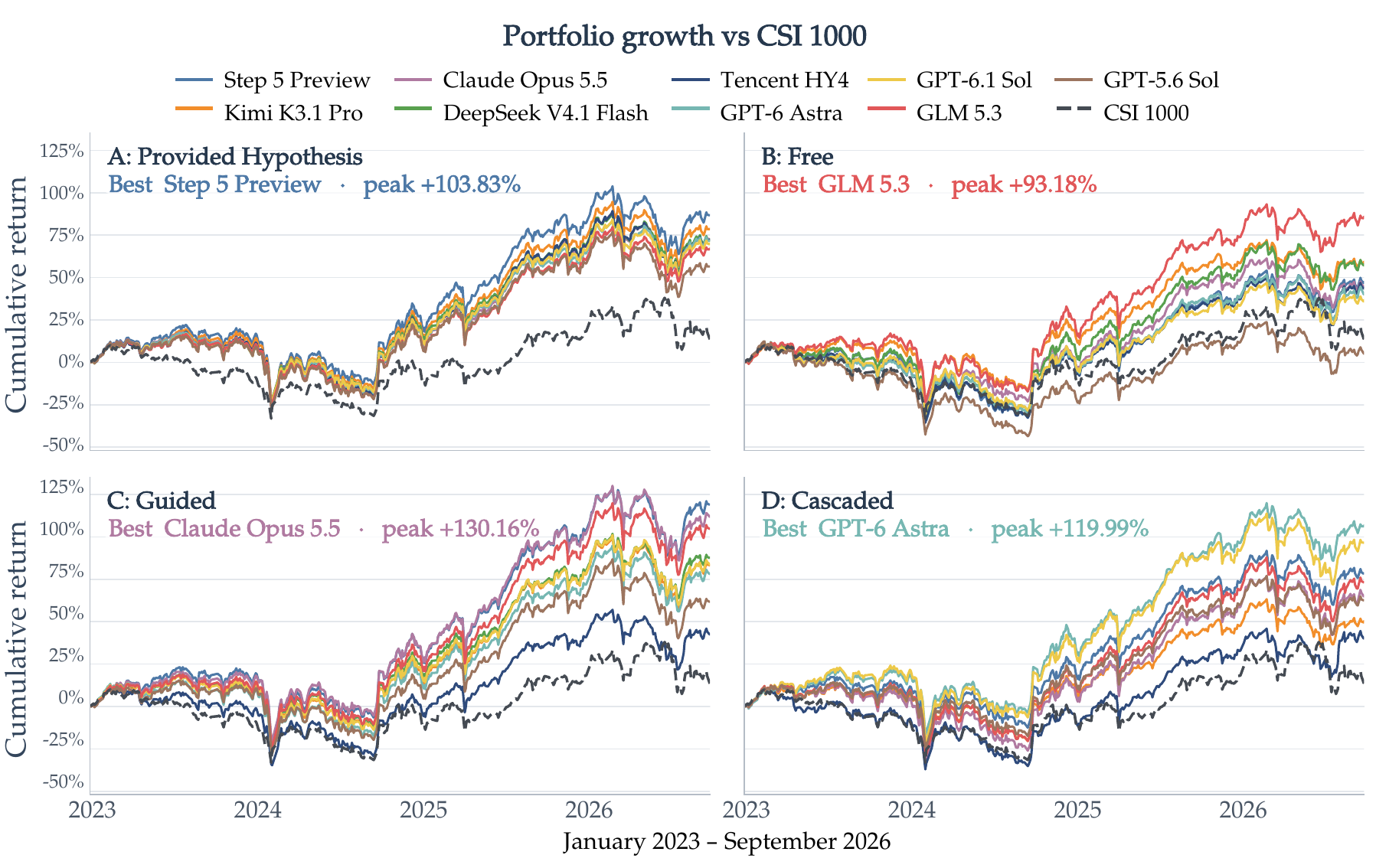}
\caption{\textbf{Cumulative portfolio returns across benchmark conditions.} Each panel shows the model portfolios, respectively, from January 2023 through September 2026, and the CSI 1000 provides the common market reference.}\label{fig:portfolio}
\end{figure}

\begin{table}[!htbp]\centering\scriptsize
\caption{\textbf{Risk and excess return of submitted-factor portfolios.} Rows follow the Provided Hypothesis ranking in Table~\ref{tab:scores}. Each cell gives the Sharpe ratio (SR), excess annualized return over the CSI 1000 index in percentage points, and signed maximum drawdown (MaxDD), in that order, for January 2023--September 2026. A maximum drawdown closer to zero indicates a smaller drawdown.}\label{tab:returns}
\setlength{\tabcolsep}{6pt}
\begin{tabular}{@{}lcccc@{}}\toprule
 & \multicolumn{4}{c}{SR\betterup{} / Excess (pp)\betterup{} / MaxDD (\%)\betterup} \\
\cmidrule(lr){2-5}
Model & Provided Hypothesis & Free & Guided & Cascaded \\ \midrule
Opus 5.5 & 0.71\,/\,+12.8\,/\,$-$37.1 & 0.54\,/\,+7.2\,/\,$-$36.1 & 0.94\,/\,+19.5\,/\,$-$33.8 & 0.63\,/\,+11.3\,/\,$-$38.6 \\
Tencent HY4 & 0.72\,/\,+12.7\,/\,$-$36.1 & 0.51\,/\,+7.1\,/\,$-$42.3 & 0.51\,/\,+6.7\,/\,$-$41.2 & 0.49\,/\,+6.3\,/\,$-$43.3 \\
GPT-6.1 Sol & 0.72\,/\,+12.3\,/\,$-$35.1 & 0.47\,/\,+5.5\,/\,$-$38.0 & 0.78\,/\,+14.8\,/\,$-$36.7 & 0.85\,/\,+17.0\,/\,$-$35.9 \\
Kimi K3.1 & 0.75\,/\,+13.9\,/\,$-$35.5 & 0.66\,/\,+10.2\,/\,$-$32.7 & 0.78\,/\,+14.7\,/\,$-$37.6 & 0.60\,/\,+8.4\,/\,$-$32.7 \\
Step 5 Preview & 0.79\,/\,+15.4\,/\,$-$36.9 & 0.53\,/\,+7.8\,/\,$-$41.2 & 0.96\,/\,+20.7\,/\,$-$36.8 & 0.78\,/\,+13.9\,/\,$-$33.0 \\
GPT-6 Astra & 0.71\,/\,+12.6\,/\,$-$36.4 & 0.49\,/\,+6.2\,/\,$-$38.6 & 0.74\,/\,+13.8\,/\,$-$36.8 & 0.95\,/\,+18.7\,/\,$-$31.5 \\
DeepSeek V4.1 & 0.73\,/\,+12.7\,/\,$-$35.3 & 0.60\,/\,+9.9\,/\,$-$39.2 & 0.81\,/\,+15.5\,/\,$-$36.7 & 
0.90\,/\,+16.2\,/\,$-$29.2 \\
GLM 5.3 & 0.68\,/\,+11.7\,/\,$-$36.3 & 0.80\,/\,+15.2\,/\,$-$34.6 & 0.89\,/\,+18.4\,/\,$-$36.1 & 0.70\,/\,+12.8\,/\,$-$37.0 \\
GPT-5.6 Sol & 0.60\,/\,+9.7\,/\,$-$36.7 & 0.19\,/\,$-$2.3\,/\,$-$49.8 & 0.64\,/\,+10.7\,/\,$-$37.6 & 0.67\,/\,+10.9\,/\,$-$34.1 \\ \bottomrule
\end{tabular}

\end{table}

\begin{finding}\obs{3}
\textbf{Cascaded broadens factor-return coverage in every paired campaign, while the additional diversity is accompanied by lower mean task quality throughout.}
\end{finding}

Relative to Guided, every Cascaded setting moves to the lower-right quadrant of Figure~\ref{fig:guidance-deltas}. Diversity increases, while mean task quality decreases. This common direction reveals a consistent quality--diversity trade-off. The response of the composite score is mixed because $S$ rewards both components. It rises in a slight majority of paired campaigns even though $Q$ falls in every comparison. DeepSeek V4.1 illustrates this distinction. Its diversity increases from 22.40 to 57.30 and its mean quality decreases from 30.06 to 20.45, yet its composite score rises from 14.23 to 15.48. Table~\ref{tab:returns} further provides a portfolio-level view of the same comparison. Relative to Guided, Cascaded improves both the Sharpe ratio and excess annualized return in four paired portfolios, while six have a maximum drawdown closer to zero. The broader factor-return coverage therefore produces mixed portfolio consequences. Return and risk-adjusted performance improve in a minority of paired campaigns, while downside risk improves more often. Taken together, the factor-level and portfolio-level results show that the additional diversity under Cascaded does not translate into a general improvement in research value. Because Cascaded also changes task organization and access to prior proposals, the results describe the combined condition rather than the isolated effect of research history.

\subsection{Discussion and Limitations}\label{sec:discussion}
Our findings support evaluating autonomous research at the problem design level. The ranking difference between Provided Hypothesis and Free indicates that execution on supplied hypotheses and autonomous hypothesis selection capture distinct capabilities. Across autonomous settings, mean quality and diversity can move in opposite directions, so the composite score alone cannot distinguish stronger average factors from broader factor-return coverage. Evaluations should therefore specify the information available during hypothesis selection and report $Q$, $D$, $S$, and portfolio outcomes separately.

The condition comparisons are descriptive because instructions, task allocation, order, and access to prior proposals can change together. Identifying their separate effects requires controlled experiments and independently randomized direction quotas and history access. The financial outcomes are specific to the evaluated market universe and period, while $D$ measures dependence among factor returns rather than semantic novelty among proposals. Independent recomputation also requires the underlying stock-by-day return matrices. The reported rankings should therefore be tested across additional model configurations, markets, and private evaluation periods.

\section{Conclusion}
AutoAlphaBench evaluates research agents in quantitative factor discovery across supplied and self-selected hypotheses. The results show that performance on supplied hypotheses does not determine rankings under autonomous selection, that direction structure is associated with stronger joint quality--diversity outcomes, and that Cascaded broadens factor-return coverage while lowering mean task quality and yielding mixed portfolio gains. These findings motivate evaluating autonomous research as a campaign-level allocation process and reporting quality, diversity, composite scores, and portfolio outcomes separately. Controlled studies of proposal assignment, direction quotas, and history access, together with replication across markets and private periods, are needed to assess causality and generalization.

\appendix
\section{Experimental Configuration}
\label{app:configuration}

\textbf{Models and agent harnesses.} Each campaign uses one model with its associated agent harness. GPT-6.1 Sol, GPT-5.6 Sol, and GPT-6 Astra run through Codex. Claude Opus 5.5, GLM 5.3, DeepSeek V4.1 Flash, Kimi K3.1 Pro, Step 5 Preview, and Tencent HY4 run through Claude Code. Harbor orchestrates the tasks in a standardized execution environment.

\begin{figure}[H]
  \centering
  \figfile[\linewidth]{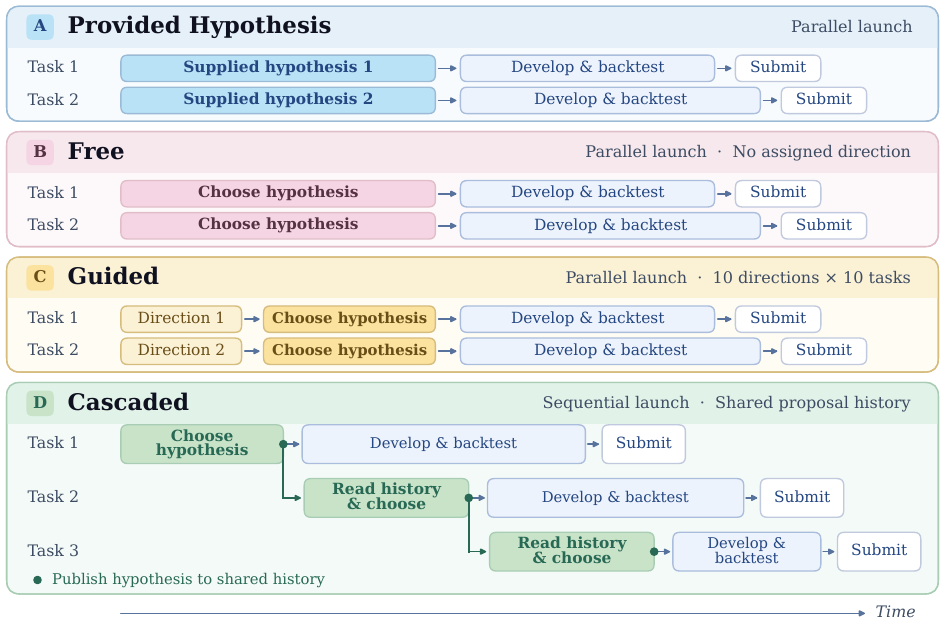}
  \caption{\textbf{Hypothesis selection and task scheduling in AutoAlphaBench.} In (A) Provided Hypothesis, the benchmark supplies a hypothesis for each task. In (B) Free, agents select hypotheses independently without an assigned direction. In (C) Guided, agents select hypotheses within their assigned research directions. Tasks in (A)--(C) start in parallel. In (D) Cascaded, each task publishes its hypothesis upon selection. The next task then starts with access to the accumulated proposal history while earlier tasks continue developing and backtesting their factors. The timelines are schematic.}
  \label{fig:appendix-session-timeline}
\end{figure}

\textbf{Session information.} Parallel tasks use isolated workspaces. Figure~\ref{fig:appendix-session-timeline} shows how hypotheses are assigned or selected and when Cascaded tasks can access earlier proposals.

\section{Development and Held-Out Diagnostics}
\label{app:extended-diagnostics}

Mean held-out Sharpe is lower than mean development Sharpe in all 26 campaigns with both summaries (Table~\ref{tab:extended-campaign}). Guided has fewer zero-quality tasks than Free for seven of the nine models.

\begin{table}[H]\centering\footnotesize
\caption{\textbf{Campaign diagnostics across the autonomous settings.} Sharpe ratios are averaged over the available observations in each period. In Panel B, \emph{Zero} counts tasks with $q_i=0$, \emph{Backtests} is the mean number of backtests per task, and \emph{Names} counts distinct recorded proposal names. Dashes indicate unavailable values.}
\label{tab:extended-campaign}
\setlength{\tabcolsep}{4pt}
\renewcommand{\arraystretch}{1.1}
\textbf{\textcolor{navy}{A. Mean Sharpe ratios}}\par\smallskip
\begin{tabular*}{\linewidth}{@{\extracolsep{\fill}}lrrrrrr@{}}
\toprule
& \multicolumn{2}{c}{Free} & \multicolumn{2}{c}{Guided} & \multicolumn{2}{c}{Cascaded} \\
\cmidrule(lr){2-3}\cmidrule(lr){4-5}\cmidrule(lr){6-7}
Model & Development & Held-out & Development & Held-out & Development & Held-out \\
\midrule
Opus 5.5 & 2.22 & $-0.06$ & 2.73 & 1.51 & 1.95 & 0.98 \\
GPT-6.1 Sol & 2.61 & 0.63 & 2.35 & 1.49 & 1.60 & 1.06 \\
GPT-5.6 Sol & 0.56 & $-0.46$ & 1.51 & 0.93 & 0.95 & 0.63 \\
GPT-6 Astra & 2.29 & 0.52 & 2.43 & 1.66 & 1.41 & 0.87 \\
GLM 5.3 & 2.42 & 1.15 & 2.49 & 1.72 & 1.57 & 0.95 \\
DeepSeek V4.1 & 2.77 & 1.77 & 2.56 & 1.58 & \NA & \NA \\
Kimi K3.1 & 1.57 & 0.67 & 2.14 & 1.42 & 1.81 & 0.97 \\
Step 5 Preview & 2.98 & 1.84 & 2.70 & 1.77 & 1.54 & 0.94 \\
HY4 & 2.70 & 1.24 & 2.64 & 1.56 & 1.93 & 1.16 \\
\bottomrule
\end{tabular*}
\par\medskip
\textbf{\textcolor{navy}{B. Zero-quality tasks, backtests, and proposal counts}}\par\smallskip
\begin{tabular*}{\linewidth}{@{\extracolsep{\fill}}lrrrrrrrrr@{}}
\toprule
& \multicolumn{3}{c}{Free} & \multicolumn{3}{c}{Guided} & \multicolumn{3}{c}{Cascaded} \\
\cmidrule(lr){2-4}\cmidrule(lr){5-7}\cmidrule(lr){8-10}
Model & Zero & Backtests & Names & Zero & Backtests & Names & Zero & Backtests & Names \\
\midrule
Opus 5.5 & 63 & 4.74 & 25 & 12 & 4.64 & 74 & 20 & 4.78 & 100 \\
GPT-6.1 Sol & 37 & 4.98 & 54 & 4 & 4.98 & 80 & 14 & 4.87 & 100 \\
GPT-5.6 Sol & 60 & 4.96 & 62 & 29 & 4.83 & 76 & 27 & 4.84 & 100 \\
GPT-6 Astra & 25 & 5.00 & 38 & 4 & 4.95 & 67 & 15 & 4.93 & 100 \\
GLM 5.3 & 11 & 4.90 & 62 & 6 & 4.90 & 84 & 16 & 4.89 & 98 \\
DeepSeek V4.1 & 5 & 4.87 & 76 & 5 & 4.75 & 86 & 11 & 4.50 & \NA \\
Kimi K3.1 & 23 & 4.96 & 52 & 6 & 4.90 & 84 & 14 & 4.86 & 100 \\
Step 5 Preview & 6 & 5.00 & 76 & 2 & 4.89 & 84 & 15 & 4.92 & 98 \\
HY4 & 14 & 4.97 & 79 & 20 & 4.72 & 70 & 13 & 5.00 & 98 \\
\bottomrule
\end{tabular*}
\end{table}

\section{Proposal and Task Diagnostics}
\subsection{Proposal Allocation in Free}
\label{app:model-profiles}
Table~\ref{tab:free-profiles} summarizes recurring research themes. The Herfindahl--Hirschman index (HHI) measures concentration across normalized proposal names using the definition in Section~\ref{sec:protocol}. Research themes are based on qualitative classification.

\begin{table}[H]\centering\footnotesize
\caption{\textbf{Proposal concentration and recurring research themes in Free.} Percentages indicate the share of each campaign's proposals classified as liquidity-related.}\label{tab:free-profiles}
\setlength{\tabcolsep}{5pt}\renewcommand{\arraystretch}{1.17}
\begin{tabularx}{\linewidth}{@{}lr>{\raggedright\arraybackslash}X@{}}
\toprule
Model & HHI & Recurring research themes \\ \midrule
Opus 5.5 & 0.112 & Liquidity (75\%), including trading activity and investor flows.\\
GPT-5.6 Sol & 0.035 & Institutional trading flows.\\
GPT-6 Astra & 0.071 & Liquidity (53\%), with recurring price reversals.\\
GPT-6.1 Sol & 0.029 & Liquidity (69\%).\\
Kimi K3.1 & 0.046 & Liquidity (50\%), with recurring trading-flow hypotheses.\\
GLM 5.3 & 0.037 & Trading flows.\\
HY4 & 0.022 & Price reversals.\\
DeepSeek V4.1 & 0.020 & Trading activity and return patterns.\\
Step 5 Preview & 0.023 & Return distributions.\\
\bottomrule\end{tabularx}\end{table}

\subsection{Illustrative Task Outcomes}
Table~\ref{tab:cases} illustrates how factors with positive development Sharpe ratios can have different held-out outcomes.

\begin{table}[H]\centering\footnotesize
\caption{\textbf{Selected task outcomes.} Sharpe ratios are the final recorded values for the development and held-out periods. $q_i$ is the held-out task quality score.}
\label{tab:cases}
\begin{tabular}{@{}llrrrr@{}}
\toprule
Setting & Research theme & Backtests & Dev.\ Sharpe & Test Sharpe & $q_i$ \\ \midrule
Free & Investor flows & 4 & 3.59 & $-0.12$ & 0.00 \\
Free & Trading activity & 5 & 1.35 & $-1.02$ & 0.00 \\
Free & Price dynamics & 5 & 4.03 & 2.38 & 47.65 \\
Guided & Volatility & 3 & 4.79 & 4.05 & 74.01 \\
Cascaded & Volatility & 5 & 4.97 & 3.93 & 72.46 \\
Cascaded & Trading flows & 5 & 4.55 & 3.58 & 67.70 \\
\bottomrule
\end{tabular}
\end{table}

\section{Candidate-Selection Sensitivity}
\label{app:k-selection}

This exploratory experiment uses development-period feedback to select one of the first $k$ candidate versions for held-out evaluation. Paired comparisons cover four models, with up to ten candidates per task and mean backtest counts of 8.4--8.7 across models. The experiment reports held-out task scores on a separate scale from the campaign composite $S$.

\begin{figure}[H]\centering\figfile[.98\linewidth]{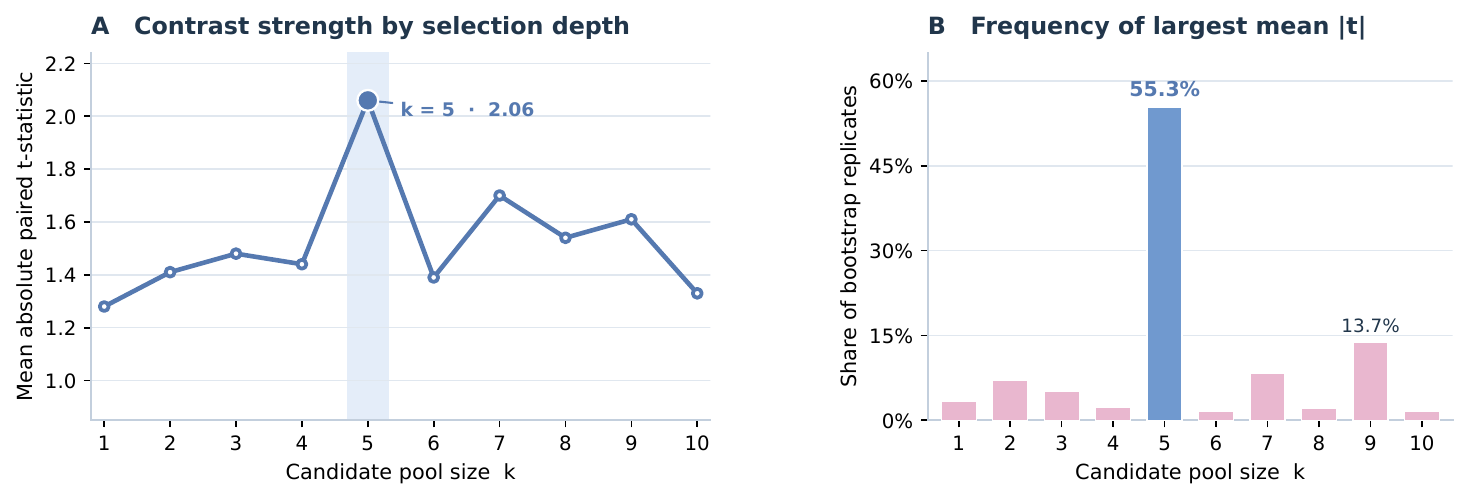}
\caption{\textbf{Sensitivity of paired model comparisons to candidate pool size.} (A) Mean absolute paired $t$-statistic across the six model pairs at each $k$. (B) Share of 2,000 bootstrap resamples in which each $k$ yielded the largest mean absolute paired $t$-statistic.}\label{fig:k-selection}
\end{figure}

Table~\ref{tab:k-selection} gives the paired task counts and results by pool size. Table~\ref{tab:k5-pairs} reports comparisons at $k=5$, the pool size with the largest observed mean absolute $t$-statistic. The pairwise $p$-values are descriptive because the pool size was selected after examining the results.

\begin{table}[H]\centering\footnotesize
\caption{\textbf{Candidate-selection results by pool size.} Panel A reports held-out scores for the four models and the number of paired tasks at each $k$. Panel B uses the same tasks and reports the range of model scores, the mean absolute paired $t$-statistic, the number of pairs with unadjusted $p<0.05$, and the share of bootstrap resamples in which $k$ maximized the mean $|t|$.}\label{tab:k-selection}
\setlength{\tabcolsep}{7pt}\renewcommand{\arraystretch}{1.14}
\textbf{A. Scores by model}\par\smallskip
\begin{tabular}{@{}rrrrrr@{}}\toprule
$k$ & Tasks & Opus 5.5\betterup & GPT-6 Astra\betterup & GLM 5.3 Flash\betterup & DeepSeek V4.1 Flash\betterup\\\midrule
1 & 78 & 19.73 & 18.25 & 16.82 & 18.31\\
2 & 97 & 20.93 & 19.70 & 18.02 & 19.55\\
3 & 98 & 21.71 & 20.37 & 18.47 & 20.84\\
4 & 98 & 23.01 & 21.62 & 19.72 & 21.99\\
\textbf{5} & 98 & 24.93 & 23.08 & 20.36 & 22.30\\
6 & 98 & 24.81 & 24.45 & 21.70 & 23.79\\
7 & 98 & 26.24 & 24.90 & 22.34 & 24.15\\
8 & 98 & 26.14 & 24.93 & 22.56 & 24.20\\
9 & 98 & 26.44 & 26.06 & 23.12 & 24.27\\
10 & 98 & 26.43 & 26.23 & 23.91 & 24.43\\
\bottomrule\end{tabular}
\par\medskip\textbf{B. Paired contrast and resampling summaries}\par\smallskip
\begin{tabular}{@{}rrrrr@{}}\toprule
$k$ & Score range & Mean $|t|$ & Significant pairs / 6 & Bootstrap share\\\midrule
1 & 2.91 & 1.28 & 1/6 & 3.3\%\\
2 & 2.91 & 1.41 & 1/6 & 7.0\%\\
3 & 3.24 & 1.48 & 1/6 & 5.1\%\\
4 & 3.29 & 1.44 & 1/6 & 2.2\%\\
\textbf{5} & 4.57 & \textbf{2.06} & 3/6 & \textbf{55.3}\%\\
6 & 3.11 & 1.39 & 2/6 & 1.6\%\\
7 & 3.90 & 1.70 & 2/6 & 8.2\%\\
8 & 3.58 & 1.54 & 2/6 & 2.1\%\\
9 & 3.31 & 1.61 & 2/6 & 13.7\%\\
10 & 2.52 & 1.33 & 1/6 & 1.5\%\\
\bottomrule\end{tabular}\end{table}

\begin{table}[H]\centering\small
\caption{\textbf{Pairwise model comparisons at $k=5$.} Mean score differences subtract the second model from the first across 98 paired tasks. The $p$-values are unadjusted.}\label{tab:k5-pairs}
\setlength{\tabcolsep}{7pt}\renewcommand{\arraystretch}{1.15}
\begin{tabularx}{\linewidth}{@{}>{\raggedright\arraybackslash}Xrrr@{}}\toprule
Comparison (first $-$ second) & Mean diff. & $t$ & $p$\\\midrule
Opus 5.5 vs. GLM 5.3 Flash & +4.57 & +3.85 & $<0.001$\\
Opus 5.5 vs. DeepSeek V4.1 Flash & +2.63 & +2.39 & 0.019\\
GPT-6 Astra vs. GLM 5.3 Flash & +2.71 & +2.14 & 0.035\\
Opus 5.5 vs. GPT-6 Astra & +1.85 & +1.79 & 0.076\\
GLM 5.3 Flash vs. DeepSeek V4.1 Flash & $-1.94$ & $-1.46$ & 0.149\\
GPT-6 Astra vs. DeepSeek V4.1 Flash & +0.78 & +0.70 & 0.482\\
\bottomrule\end{tabularx}\end{table}

\section{Scope and Evidence Tiers}
\textbf{Recorded outcomes.} The records contain the reported benchmark scores, $Q$ and $D$ where applicable, task counts, development and test Sharpe ratios, proposal names, delivery and audit status, and selected tool events. \textbf{Descriptive analyses.} The analysis compares proposal concentration, held-out quality, and factor diversity across the available model runs and settings. \textbf{Questions for controlled experiments.} The effects of access to prior proposals, task quotas, development feedback, and possible training-data exposure require separate tests.


\begin{thebibliography}{20}\small
\bibitem[Mo et al.(2026)]{agentideabench} Y. Mo et al. (2026). \emph{AgentIdeaBench: Benchmarking Scientific Ideation in the Agent Era}. arXiv:2609.07611. \url{https://arxiv.org/abs/2609.07611}.
\bibitem[Xu et al.(2026)]{researchclawbench} W. Xu et al. (2026). \emph{ResearchClawBench: A Benchmark for End-to-End Autonomous Scientific Research}. arXiv:2606.07591. \url{https://arxiv.org/abs/2606.07591}.
\bibitem[Qian et al.(2026)]{quantarena} H. Qian et al. (2026). \emph{Quant-Arena: Benchmarking LLM Agents for Quantitative Trading Research}. arXiv:2610.30113. \url{https://arxiv.org/abs/2610.30113}.
\bibitem[Tang et al.(2025)]{alphaagent} Z. Tang et al. (2025). \emph{AlphaAgent: LLM-Driven Alpha Mining with Regularized Exploration to Counteract Alpha Decay}. KDD. \url{https://arxiv.org/abs/2502.16789}.
\bibitem[Han et al.(2026)]{quantaalpha} J. Han et al. (2026). \emph{QuantaAlpha: An Evolutionary Framework for LLM-Driven Alpha Mining}. arXiv:2602.07085. \url{https://arxiv.org/abs/2602.07085}.
\bibitem[Zhang et al.(2026a)]{alphaforge} W. Zhang et al. (2026a). \emph{AlphaForgeBench: Benchmarking End-to-End Trading Strategy Design with Large Language Models}. arXiv:2602.18481. \url{https://arxiv.org/abs/2602.18481}.
\bibitem[Luo et al.(2026)]{alphabench} H. Luo et al. (2026). \emph{AlphaBench: Benchmarking Large Language Models in Formulaic Alpha Factor Mining}. ICLR. \url{https://proceedings.iclr.cc/paper_files/paper/2026/hash/4f3820576130a8f796ddbf204c841487-Abstract-Conference.html}.
\bibitem[Bailey et al.(2017)]{bailey} D. H. Bailey, J. M. Borwein, M. L\'opez de Prado, and Q. J. Zhu. (2017). The probability of backtest overfitting. \emph{Journal of Computational Finance}, 20(4), 39--69.
\bibitem[Lu et al.(2026)]{lu2026aiscientist} C. Lu et al. (2026). \emph{Towards end-to-end automation of AI research}. Nature, 651, 914--919. \url{https://doi.org/10.1038/s41586-026-10265-5}.
\bibitem[Li et al.(2025)]{li2025rdagentquant} Y. Li et al. (2025). \emph{R\&D-Agent-Quant: A Multi-Agent Framework for Data-Centric Factors and Model Joint Optimization}. NeurIPS. \url{https://doi.org/10.52202/085713-3967}.
\bibitem[Si et al.(2025)]{si2025ideas} C. Si, D. Yang, and T. Hashimoto. (2025). \emph{Can LLMs Generate Novel Research Ideas? A Large-Scale Human Study with 100+ NLP Researchers}. ICLR. \url{https://arxiv.org/abs/2409.04109}.
\bibitem[Wang et al.(2024)]{wang2024scimon} Q. Wang, D. Downey, H. Ji, and T. Hope. (2024). \emph{SciMON: Scientific Inspiration Machines Optimized for Novelty}. ACL. \url{https://aclanthology.org/2024.acl-long.18/}.
\bibitem[Starace et al.(2025)]{starace2025paperbench} G. Starace et al. (2025). \emph{PaperBench: Evaluating AI's Ability to Replicate AI Research}. ICML. \url{https://proceedings.mlr.press/v267/starace25a.html}.
\bibitem[Wijk et al.(2025)]{wijk2025rebench} H. Wijk et al. (2025). \emph{RE-Bench: Evaluating Frontier AI R\&D Capabilities of Language Model Agents against Human Experts}. ICML. \url{https://proceedings.mlr.press/v267/wijk25a.html}.
\bibitem[Yu et al.(2023)]{yu2023alphagen} S. Yu et al. (2023). \emph{Generating Synergistic Formulaic Alpha Collections via Reinforcement Learning}. KDD. \url{https://doi.org/10.1145/3580305.3599831}.
\bibitem[Wang et al.(2025)]{wang2025alphagpt} S. Wang et al. (2025). \emph{Alpha-GPT: Human-AI Interactive Alpha Mining for Quantitative Investment}. EMNLP System Demonstrations. \url{https://aclanthology.org/2025.emnlp-demos.14/}.
\bibitem[Wang et al.(2026)]{wang2026factorminer} Y. Wang et al. (2026). \emph{FactorMiner: A Self-Evolving Agent with Skills and Experience Memory for Financial Alpha Discovery}. arXiv:2602.14670. \url{https://arxiv.org/abs/2602.14670}.
\bibitem[Zhang et al.(2026b)]{report2alpha} Y. Zhang, H. Yang, and W. Meng. (2026b). \emph{Report2Alpha-Bench: Evaluating LLM Agents on End-to-End Quantitative Research}. arXiv:2611.27635. \url{https://arxiv.org/abs/2611.27635}.
\bibitem[Harvey et al.(2016)]{harvey2016factorzoo} C. R. Harvey, Y. Liu, and H. Zhu. (2016). \emph{\ldots and the Cross-Section of Expected Returns}. Review of Financial Studies, 29(1), 5--68. \url{https://doi.org/10.1093/rfs/hhv059}.
\bibitem[Huang et al.(2024)]{huang2024mlagentbench} Q. Huang, J. Vora, P. Liang, and J. Leskovec. (2024). \emph{MLAgentBench: Evaluating Language Agents on Machine Learning Experimentation}. ICML. \url{https://proceedings.mlr.press/v235/huang24y.html}.
\bibitem[Doshi and Hauser(2024)]{doshi2024diversity} A. R. Doshi and O. P. Hauser. (2024). \emph{Generative AI enhances individual creativity but reduces the collective diversity of novel content}. Science Advances, 10(28), eadn5290. \url{https://doi.org/10.1126/sciadv.adn5290}.
\bibitem[Hou et al.(2020)]{hou2020anomalies} K. Hou, C. Xue, and L. Zhang. (2020). \emph{Replicating Anomalies}. Review of Financial Studies, 33(5), 2019--2133. \url{https://doi.org/10.1093/rfs/hhy131}.
\end{thebibliography}
\end{document}